\documentclass{article} 
\usepackage{times}
\usepackage{graphicx} 
\usepackage{subfigure}

\usepackage{natbib}

\usepackage{algorithm}
\usepackage{algorithmic}

\usepackage{hyperref}

\usepackage[accepted]{icml2017}

\icmltitlerunning{Loss is its own Reward}

\usepackage{amsmath, amsfonts, amssymb, amsthm}
\usepackage{booktabs}
\usepackage{soul}

\newcommand{\E}{\mathbb{E}}
\newcommand{\Sc}{\mathcal{S}}
\newcommand{\Ac}{\mathcal{A}}
\newcommand{\Rb}{\mathbb{R}}

\newcommand{\minisection}[1]{\textbf{#1}\hspace{0.3em}}

\begin{document}

\twocolumn[
\icmltitle{Loss is its own Reward:\\Self-Supervision for Reinforcement Learning}

\icmlsetsymbol{equal}{*}

\begin{icmlauthorlist}
\icmlauthor{Evan Shelhamer}{ucb,openai}
\icmlauthor{Parsa Mahmoudieh}{ucb}
\icmlauthor{Max Argus}{ucb}
\icmlauthor{Trevor Darrell}{ucb}
\end{icmlauthorlist}

\icmlaffiliation{ucb}{UC Berkeley}
\icmlaffiliation{openai}{OpenAI}
\icmlcorrespondingauthor{Evan Shelhamer}{shelhamer@cs.berkeley.edu}

\icmlkeywords{reinforcement learning, self-supervision, deep learning}

\vskip 0.3in
]

\printAffiliationsAndNotice{}  

\begin{abstract}
  Reinforcement learning optimizes policies for expected cumulative reward.
  Need the supervision be so narrow?
  Reward is delayed and sparse for many tasks, making it a difficult and impoverished signal for end-to-end optimization.
  To augment reward, we consider a range of self-supervised tasks that incorporate states, actions, and successors to provide auxiliary losses.
  These losses offer ubiquitous and instantaneous supervision for representation learning even in the absence of reward.
  While current results show that learning from reward alone is feasible, pure reinforcement learning methods are constrained by computational and data efficiency issues that can be remedied by auxiliary losses.
  Self-supervised pre-training and joint optimization improve the data efficiency and policy returns of end-to-end reinforcement learning.
\end{abstract}

\section{Introduction}

End-to-end reinforcement learning (RL) addresses representation learning at the same time as policy optimization.
Of these dual pursuits, current work focuses on the reinforcement learning aspects of the problem such as stochastic optimization and exploration.
Once a loss on reward is defined the representation is delegated to backpropagation without further attention to other supervisory signals.
We argue that representation learning is a bottleneck in current approaches bound by reward.
Our self-supervised auxiliary losses broaden the horizons of reinforcement learning agents to learn from all experience, whether rewarded or not.

\begin{figure*}[h]
\begin{center}
   \includegraphics[width=1.2\textwidth]{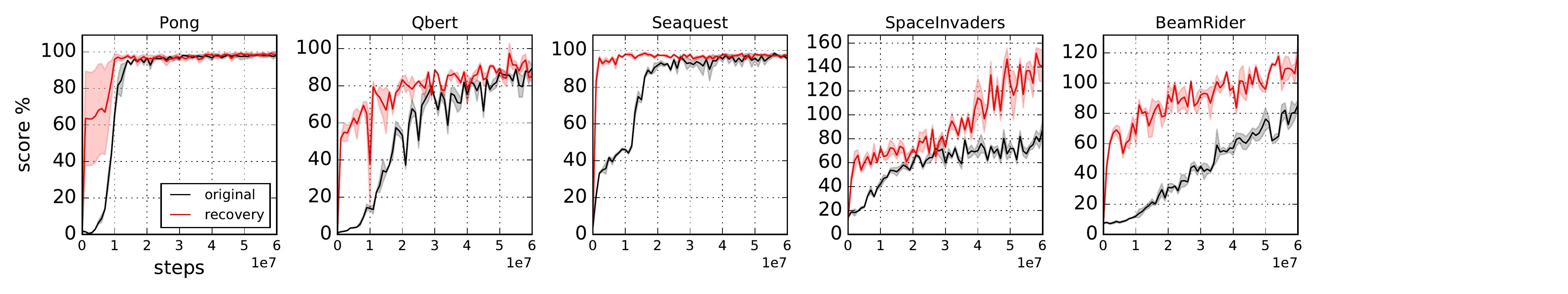}
\end{center}
\caption{
  Current methods require many transitions to arrive at good policies, but policies are often quickly recovered from their representation.
  To separate reinforcement learning from representation learning, we decapitate the agent by destroying its policy and value output parameters, and then re-train end-to-end.
  Although the policy distribution and value estimates are obliterated, most of the parameters are preserved and the policy is swiftly recovered.
  The gap between the initial optimization and recovery illustrates a representation learning bottleneck.
}
\label{fig:polrec}
\end{figure*}

To illustrate the critical role of representation learning, we show that re-training a decapitated agent, having destroyed the policy and value outputs while preserving the rest of the representation, is far faster than the initial training (Figure \ref{fig:polrec}).
Although the policy distribution and value function are lost, they are readily recovered given a representation from RL, even though the optimization and exploration issues remain.
With the importance of representation established, we turn to self-supervision to take an ambient approach to RL attuned to reward and environment alike.

Self-supervision defines losses via surrogate annotations that are synthesized from bare, unlabeled inputs.
In the context of RL, reward captures the task while self-supervision captures the environment.
In this setting, every transition contributes gradients of ambient environmental signals.
While reward might be delayed and sparse, the losses from self-supervision are instantaneous and ubiquitous.
Augmenting RL with these auxiliary losses enriches the representation through multi-task learning and improves policy optimization.

We concentrate on auxiliary losses for state, dynamics, inverse dynamics, and reward that can be formulated in a discriminative fashion.
To help RL, we transfer the representation from self-supervised pre-training with these losses.
In the other direction, we inspect the contents of policy representations by examining transfer from RL to self-supervised tasks.
Pre-training for Atari reaches higher returns with better data efficiency for a $1.4\times$ speed-up on average to $95\%$ of the best return.
Joint optimization improves further still.

\section{Preliminaries}

We briefly review policy gradient methods for RL and then frame self-supervised learning and relate it to supervised and unsupervised learning.

\subsection{Reinforcement Learning}
Reinforcement learning (RL) is concerned with policy optimization on Markov decision processes (MDPs).

Consider an MDP defined by the tuple $(\Sc, \Ac, T, R, \gamma)$, where $\Sc$ is the set of states, $\Ac$ is the set of actions, $T :\Sc \times \Ac \times \Sc \to [0, 1]$ is the transition probability distribution, $R :\Sc \times \Ac \times \Sc \to \Rb$ is the reward function, and $\gamma \in (0, 1)$ is the discount.
In addition let $p_0$ be the distribution of the initial state $s_0$.

Let $\pi$ be a stochastic policy $\pi : \Sc \times \Ac \to [0, 1]$, and $\pi_\theta$ be a policy parameterized by $\theta$.

The objective is to maximize the expected return $\eta(\pi)$ of the policy:
\begin{align*}
  \eta(\pi) = \E_{s_0,a_0,\dots}\left[\sum_{t=0}^\infty \gamma^t r(s_t, a_t, s_{t+1})\right], \text{where}\\
  s_0 \sim p_0(s_0), \; a_t \sim \pi(a_t | s_t), \; s_{t+1} \sim T(s_{t+1} | s_t, a_t)
\end{align*}
The expected return is measured by the state-action value $Q_\pi$, the value $V_\pi$, and the advantage $A_\pi$:
\begin{align*}
  Q_\pi(s_t, a_t) &= \E_{s_{t+1}, a_{t+1}, \dots}\left[\sum_{t=0}^\infty \gamma^t r(s_t, a_t, s_{t+1})\right], \\
  V_\pi(s_t)      &= \E_{a_t, s_{t+1}, \dots}\left[\sum_{t=0}^\infty \gamma^t r(s_t, a_t, s_{t+1})\right], \\
  \text{and } A_\pi(s, a)     &= Q_\pi(s,a) - V_\pi(s)
\end{align*}
where $a_t \sim \pi(a_t | s_t)$ and $s_{t+1} \sim T(s_{t+1} | s_t, a_t)$.

Policy gradient methods iteratively optimize the policy return by estimating the gradient of the expected return with respect to the policy parameters
\begin{equation*}
  \nabla_\theta \E\left[\sum_{t=0}^\infty \gamma^t r_t\right] = \E\left[\sum_{t=0}^\infty \gamma^t r_t \nabla_\theta \log \pi_\theta(a_t | s_t) \right], \\
\end{equation*}
where the expectation is sampled by executing the policy in the environment.
To improve optimization, in an actor-critic method the policy gradient can be scaled not by the return itself but by an estimate of the advantage \cite{sutton1998reinforcement}.

In this work we augment the policy gradient with auxiliary gradients from self-supervised tasks.

\subsection{Self-Supervision}

End-to-end RL admits policy learning in lieu of policy design in much the same way that end-to-end supervised learning has seen the advance of feature learning over feature design.
Supervised learning, especially as carried out for computer vision, has recently seen the rise of deeper and higher-capacity networks trained by backpropagation, reaching $100+$ layers \citep{he2016deep}.
These capacities are sustained only by massive amounts of annotation and other supervisory signals.
Supervised pre-training on large-scale annotations as exemplified by ImageNet \citep{deng2009imagenet} currently delivers the most effective features for transfer learning to other tasks.
However, a wave of renewed interest in unsupervised and ``self-supervised'' learning offers alternatives that we catalogue here \citep{doersch2015unsupervised, noroozi2016unsupervised, zhang2016colorful, donahue2016adversarial}.

To illustrate the differences, consider three kinds of learning by their objectives:

\begin{itemize}
  \item supervised learning $\min_\theta \E\left[L_\text{dis}(f_\theta(x), y)\right]$
  \item unsupervised learning $\min_\theta \E\left[L_\text{gen}(f_\theta(x), x)\right]$
  \item self-supervised learning $\min_\theta \E\left[L_\text{dis}(f_\theta(x), s(x))\right]$ with surrogate annotation function $s(\cdot)$
\end{itemize}

for data $x$, annotations $y$, losses $L$ either discriminative or generative, and parametric model $f_\theta$.
Both unsupervised learning and self-supervised learning define losses without annotation, but unsupervised learning has historically focused on generative or reconstructive losses, while nascent self-supervised methods instead define surrogate losses and synthesize the annotations from the data.
Since self-supervised and unsupervised methods can make use of unannotated data, as auxiliary losses for RL they promise to mine more from the data already available to the policy.

\begin{figure*}[t]
\begin{center}
   \includegraphics[width=\linewidth]{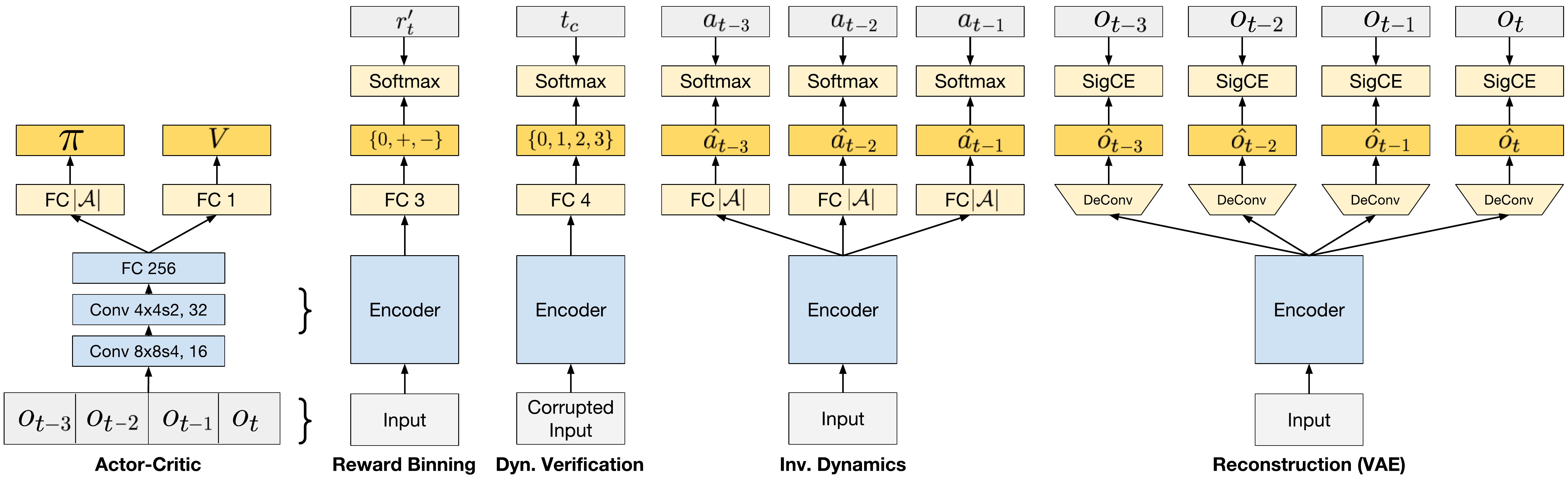}
\end{center}
\caption{
  Architectures for reinforcement learning and self-supervision.
  The actor-critic architecture is based on A3C \citep{mnih2015human} but with capacity reduced for experimental efficiency.
  The self-supervised architectures share the same encoder as the actor-critic for transferability.
  Each self-supervised task augments the architecture with its own decoder and loss.
}
\label{fig:tasks}
\end{figure*}

\section{Self-Supervision of Policies}

The state, action, reward, and successor $(s, a, r, s')$ transition standard to RL admits many kinds of self-supervision.
We explore the use of surrogate annotations that span different parts of the transitions to gauge what is informative for RL.
These diverse, ambient losses mine further supervision from the same data available to existing RL methods.

Adopting self-supervision for RL raises issues of multi-task optimization and statistical dependence.
Policy optimization and self-supervision may need to be reconciled to learn from both reward and auxiliary losses without interference.
As for the data distribution, in the RL setting the distribution of transitions is neither i.i.d. nor stationary, so self-supervision should follow the policy distribution.
We first take the simple approach of self-supervised pre-training followed by pure RL.
For pre-training we only optimize auxiliary losses on the initial, random policy distribution and do not track the policy distribution.
To remedy this and achieve further gains we switch to joint optimization of reinforcement learning and self-supervision.

\subsection{Tasks}
\label{sec:tasks}

For RL transfer, the self-supervised tasks must make use of the same transition data as RL while respecting architectural compatibility with the agent network.
We first survey auxiliary losses and then define their instantiations for our chosen environment and architecture.
Every self-supervised task augments a common, agent-compatible encoder with a task-specific decoder.
Once pre-training is complete the decoder is discarded and the shared representation is transferred to the initial agent network.

Figure \ref{fig:tasks} illustrates tasks and architectures.

\minisection{Reward}
Self-supervision of reward is a natural choice to tune the representation for RL.
Reward can be cast into a proxy task as instantaneous prediction by regression or binning into positive, zero, and negative classes.
Our self-supervised reward task is to bin $r_t$ into $r'_t \in \{0, +, -\}$ with equal balancing of the classes as done independently by \citet{jaderberg2016reinforcement}.
This is equivalent to one-step or zero-discount value function estimation, and so may seem redundant for value methods.
However, the gradient of the instantaneous prediction task is less noisy because it is not subject to policy stochasticity or bootstrapping error.
With reward, the proxy task accuracy is expected to closely mirror the degree of policy improvement.

\minisection{Dynamics and Inverse Dynamics}
Surrogate annotations for these tasks capture state, action, and successor $(s, a, s')$ relationships from transitions.
Even a single transition suffices to define losses on dynamics (successors) and inverse dynamics (actions).
The losses need not form a transition model, and simple proxies can suffice to help tune the representation.
The difficulty of temporal self-supervision can be adjusted through the span and stride of time steps.

Dynamics can be cast into a verification task by recognizing whether state-successor $(s, s')$ pairs are drawn from the environment or not.
This can be made action conditional by extending the data to $(s, a, s')$ and solving the same classification task.
Our self-supervised dynamics verification task is to identify the corrupted observation $o_{t_c}$ in a history from $t_0$ to $t_k$, where $o_{t_c}$ is corrupted by swapping it with $o_{t'}$ for $t' \notin \{t_0, \dots, t_k\}$.
We synthesize negatives by transplanting successors from other, nearby time steps.
While the transition function is not necessarily one-to-one, and the synthetic negatives are noisy, in expectation these surrogate annotations will match the transition statistics.

Inverse dynamics, mapping $\Sc \times \Sc \to \Ac$, can be reduced to classification (for discrete actions) or regression (for continuous actions).
Our self-supervised inverse dynamics task is to infer the intervening actions of a history of observations.
When $|\Ac| << |\Sc|$, as is often the case, the self-supervision of inverse dynamics may be more statistically and computationally tractable.

\minisection{Reconstruction}
Auto-encoding/AE \cite{hinton2006reducing} and variational auto-encoding/VAE \cite{kingma2014auto-encoding,rezende2014stochastic} learn to reconstruct the input subject to a representational bottleneck.
Generative adversarial networks/GANs \cite{goodfellow2014generative} optimize a generator and discriminator to learn a model of the data, to which bidirectional GANs \cite{donahue2016adversarial} add an encoder for adversarial feature learning.
The surrogate annotation for reconstruction is simply the identity as the loss is a distance between the input and output.
While a popular line of attack for unsupervised learning, the representations learned by reconstruction are relatively poor for transfer \citep{donahue2016adversarial}.
Nevertheless we include reconstruction for comparison with self-supervised tasks that map inputs to distinct surrogate annotations.

\minisection{Observation Cues}
A number of visual signals have been identified that help learn transferable features.
Visual coherence and context \citep{doersch2015unsupervised, noroozi2016unsupervised, pathak2016context}  are cast into losses by discriminatively recognizing spatial relationships (as in solving a jigsaw puzzle) or generating input pixels (as in in-painting).
Colorization of greyscale imagery \citep{zhang2016colorful} or more generally any image-to-image mapping between modalities can be cast into pixelwise auxiliary losses.
As the policy acts across transitions, and dependence spans time, it may be insufficient to self-supervise observations alone.
In preliminary experiments these losses had no effect so we do not pursue them further.

In our approach the purpose of self-supervision is representation learning and not full modeling of the dynamics and reward.
As illustrated by these proxy tasks, the surrogate annotations need not directly predict the transition and reward functions.
The auxiliary losses are expected to give gradients and not necessarily furnish a generative model for model-based RL.
While modeling could be intractable, the gradients might suffice to improve reinforcement learning.

\subsection{Loss as Intrinsic Reward}

Intrinsic rewards are intended to scaffold skill learning, aid exploration, or otherwise guide the policy to improve \citep{barto2004intrinsically, chentanez2004intrinsically}.
Rewards that formalize novelty, curiosity, and competence focus on learning progress and predictive error \citep{schmidhuber1991curious, oudeyer2005playground, houthooft2016variational}.
Self-supervisory losses could serve as intrinsic rewards of this kind, and simultaneously guide the policy while tuning the representation through gradients.

Self-supervisory intrinsic rewards could lead the policy to novel and unlearned states for exploration.
Following the loss could visit the transitions to still be learned, until they learned, and then move on.
It may be crucial to reward learning progress, and not the absolute loss, to ensure improvement.
This in effect importance samples by the auxiliary losses.
A baseline for this directed pre-training is self-supervision on the static data distribution of a fixed random policy.
Unifying loss and reward in this way is an underexplored opportunity supplied by end-to-end RL.

\section{Results}

We show results on self-supervision for policy pre-training and joint optimization on Atari.
To begin we check the feasibility of the self-supervised tasks on transitions collected from random policies.
Then for each proxy task and environment we measure improvements in return and data efficiency for self-supervised policy pre-training.
As a probe into policy representations, we examine decoding from fixed reinforcement learning weights to proxy tasks.
Policies trained with self-supervision converge to the same or better return and do so in fewer updates.

\subsection{Self-Supervision}

\begin{table}[t]
\centering
\scalebox{0.64}{
\begin{tabular}{rrrrrrr}
\toprule
& Pong & Qbert & Seaquest & S. Invaders & BeamRider & Breakout \\
\midrule
VAE [$\ell_2$]      & 1.6          & 1.8          & 2.5         & 1.7          & 2.5          & 1.2 \\
BiGAN [$\ell_2$]    & 4.5          & 5.7          & 6.1         & 6.6          & 11.8         & 5.8 \\
\dots obs. mode     & 2.48         & 8.34         & 8.00        & 16.13        & 59.7         & 14.4 \\
Reward [F1]         & 0.99         & 0.82         & \st{0.03}   & 0.38         & 0.16         & 0.90 \\
Dyn. Ver. [acc. \%] & 97.5         & 92.8         & 95.0        & 90.5         & 98.6         & 70.8 \\
\dots chance        & 25           & $\cdots$     & $\cdots$    & $\cdots$     & $\cdots$     & $\cdots$ \\
Inv. Dyn. [acc. \%] & 34.9         & 17.5         & 25.5        & 33.3         & 21.1         & 33.9 \\
\dots chance        & $16.\bar{6}$ & $16.\bar{6}$ & $5.\bar{5}$ & $16.\bar{6}$ & $11.\bar{1}$ & $16.\bar{6}$ \\
\bottomrule
\end{tabular}
}
\caption{
Feasibility of the self-supervised tasks for Atari.
Most tasks reach reasonable performance.
Task metrics improve through training and optimization converges quickly in less than ten epochs.
}
\label{tab:selfsup}
\end{table}

\begin{figure*}[t!]
\begin{center}
   \includegraphics[width=\linewidth]{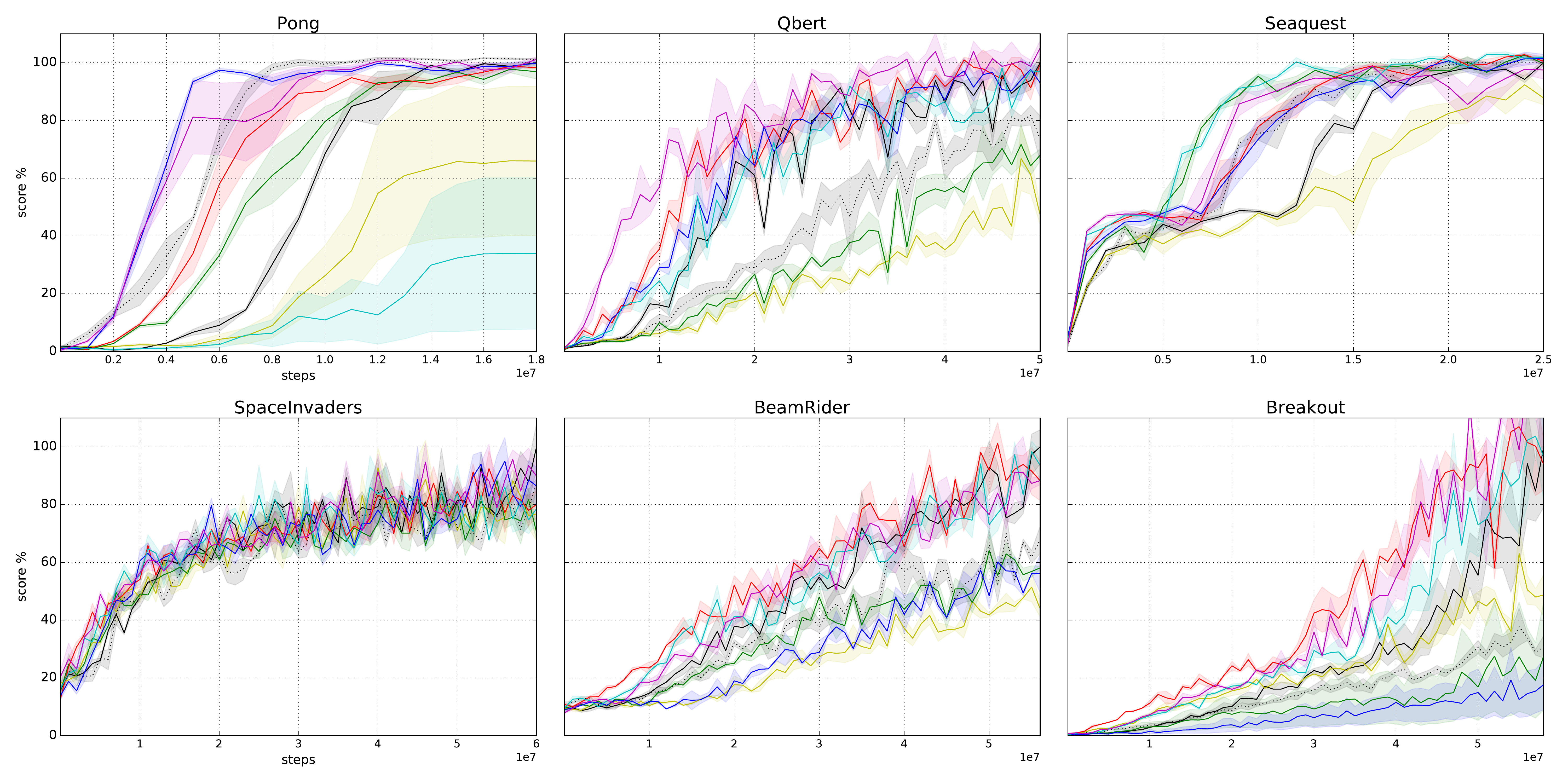}
   \includegraphics[width=0.97\linewidth]{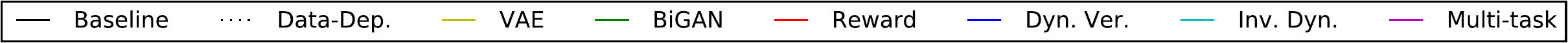}
\end{center}
\caption{
  Optimization progress for reinforcement learning from self-supervised pre-training.
  Progress is reported as percentage of the best baseline return with evaluation every 1M updates.
  Many self-supervised tasks improve data efficiency without sacrificing return.
  Tasks independent of reward, such as dynamics and inverse dynamics, can nevertheless improve optimization.
  Improvement is strongest early in training when the pre-training and policy distributions are close.
  Refer to Section \ref{sec:tasks} for the details of the auxiliary tasks.
  The mean and variance of the score is calculated over three runs.
}
\label{fig:polopt}
\end{figure*}

Our collection of self-supervised policies for Atari are variations of the asynchronous advantage actor-critic (A3C) architecture of \citet{mnih2016asynchronous}.
The actor-critic network is taken as an encoder to which each task attaches its own decoder.
To begin, we optimize the proxy tasks for their own sake to check their admissibility as pre-training for RL.
In general, the self-supervised tasks achieve reasonable performance and converge quickly.
The task metrics across several environments are reported in Table \ref{tab:selfsup}.
Note that proxy task performance need not be perfect to yield a transferable representation, and indeed low proxy task accuracy can still deliver state-of-the-art self-supervised features \cite{doersch2015unsupervised}.

Multi-task training of the reward, dynamics, and inverse dynamics tasks achieves comparable scores.
It was not necessary to balance losses or otherwise tune learning for multi-task optimization of these auxiliary losses.
The encoder apparently has enough capacity to jointly address these proxy tasks.
A higher-capacity encoder may do better still, and the interaction of encoder capacity, self-supervision, and policy optimization could compound gains for deeper architectures in richer environments.

\subsection{Policy Pre-training}

Self-supervised pre-training followed by RL fine-tuning is the simplest approach to incorporating auxiliary losses.
This simplicity controls for confounds in joint optimization such as loss weighting and learning rate schedules.
Any effect of self-supervision is purely due to representation learning prior to reinforcement learning.

\begin{figure*}[t]
\begin{center}
   \includegraphics[width=\linewidth]{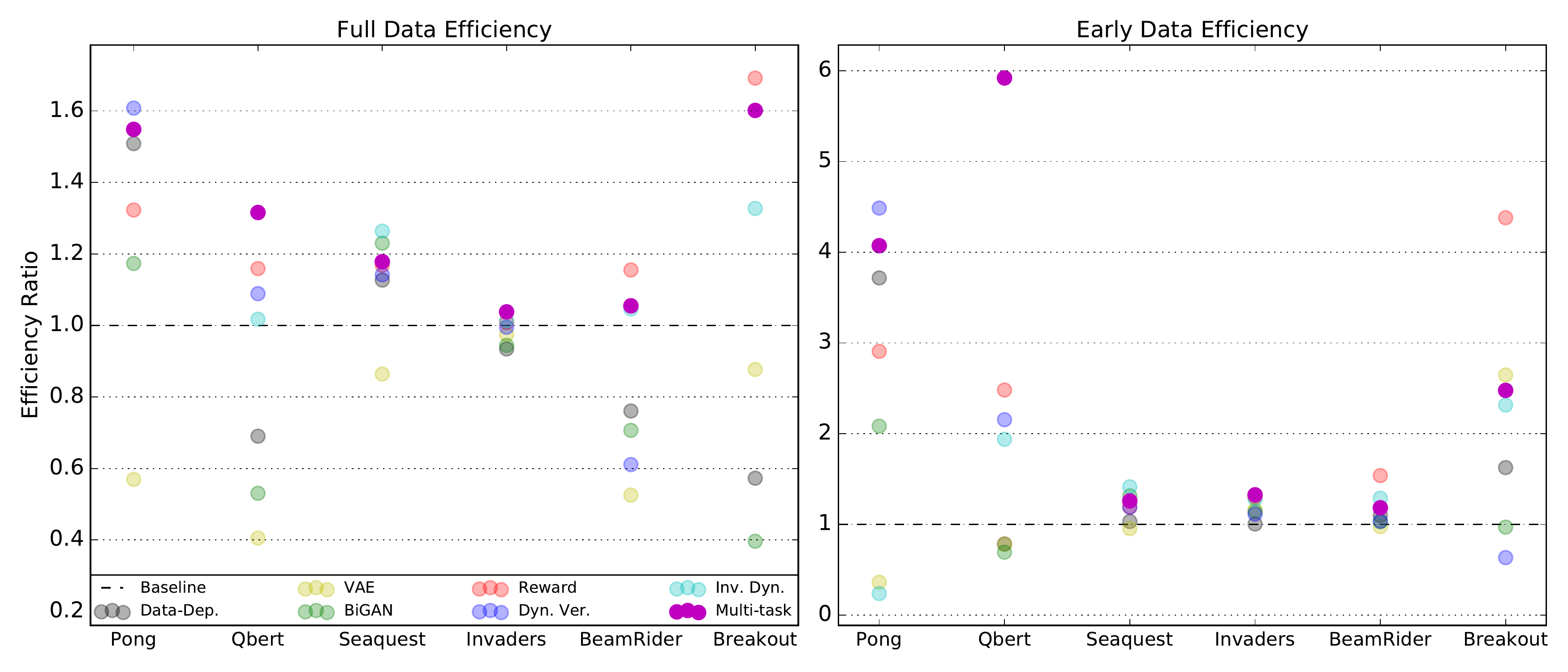}
\end{center}
\vspace{-5mm}
\caption{
  Data efficiency of RL with self-supervised pre-training.
  To measure data efficiency, we calculate the area under the score/iteration curve and report the ratio to the baseline.
  Multi-task self-supervision improves $1.3\times$ on average for full optimization to 60M iterations.
  Focusing on early optimization,  multi-task self-supervision gives $2.7\times$ improvement for the first 10M iterations.
}
\label{fig:poltime}
\end{figure*}

We compare simple initialization strategies---random initialization as well as calibrated and data-dependent initialization \citep{krahenbuhl2015data}---with our self-supervised tasks.
The calibrated random initialization has little effect while data-dependent initialization variably helps and hurts.
Our self-supervised tasks boost RL further and do so in more cases.
These tasks include auxiliary losses that are agnostic to reward, letting learning make progress without it.
Perhaps surprisingly, self-supervision of reward is not universally the most effective pre-training.
Figure \ref{fig:polopt} shows policy optimization progress with the various pre-training schemes, Figure \ref{fig:poltime} reports data efficiency, and Table \ref{tab:polret} reports policy returns.

The immediate observation is that pre-training suffices to improve optimization.
Returns at convergence are nearly equal or better than baseline and the optimization is more data efficient.
The sole exception is when pre-training diverges, but this is simple to diagnose.
Pre-training is most helpful early in the optimization, when the policy distribution is close to the random distribution (which is the data distribution for pre-training).
In the few-shot or budgeted regimes, there is a steeper advantage to self-supervision.

Overall the self-supervised tasks surpass reconstructive tasks.
Reconstruction by VAE is mostly harmful, but on the other hand BiGAN results show some improvement.
However, there is no clear ordering of the individual tasks across environments, neither for policy return nor for data efficiency.
Multi-task optimization of reward, dynamics, and inverse dynamics tends to improve on both fronts.
When ranked by data efficiency, the median rank of the baseline across environments is 4.5 (out of 8) while multi-task pre-training ranks second.
Multi-task self-supervision is a practical default.

\begin{table}[t]
\centering
\scalebox{0.69}{
\begin{tabular}{rrrrrrr}
\toprule
& Pong & Qbert & Seaquest & S. Invaders & BeamRider & Breakout \\
\midrule
Baseline  & $21$  & $18028$  & $1756$  & $1102$  & $5061$  & $367$ \\
\midrule
Data-Dep.  & $100\%$ & $90\%$  & $100\%$ & $99\%$  & $74\%$ & $39\%$ \\
VAE        & $100\%$ & $82\%$  & $99\%$  & $\mathbf{107}\%$ & $51\%$ & $62\%$ \\
BiGAN      & $101\%$  & $84\%$  & $100\%$  & $83\%$  & $61\%$  & $1\%$ \\
Reward     & $100\%$ & $101\%$ & $100\%$ & $91\%$  & $97\%$ & $96\%$ \\
Dyn. Ver.  & $99\%$  & $105\%$ & $101\%$ & $102\%$ & $61\%$ & $37\%$ \\
Inv. Dyn.  & $100\%$ & $97\%$  & $101\%$ & $100\%$ & $96\%$ & $102\%$ \\
Multi-task & $\mathbf{101}\%$ & $\mathbf{111}\%$ & $\mathbf{102}\%$ & $105\%$ & $\mathbf{99}\%$ & $\mathbf{110}\%$ \\
\bottomrule
\end{tabular}
}
\caption{
  Returns from by self-supervised pre-training.
  Returns are reported as the absolute return for the baseline (pure RL from random initialization) and the return relative to the baseline for the other conditions.
  The returns achieved are nearly equal or better.
}
\label{tab:polret}
\end{table}

\subsection{Probing Policy Representations}

Transfer from self-supervision to RL scaffolds the policy representation and improves optimization.
However, whether transfer helps by capturing aspects of the environment or merely conditioning the weights is unclear.
Furthermore, it is not obvious what is encoded by policy representations learned by pure RL.

To gather indirect evidence, we explore transfer from RL to our proxy tasks to see which can be decoded from fixed parameters.
For each task we affix a decoder to the feature layer from which policy and value predictions are made.
The decoder is learned and evaluated on data from the policy distribution at the end of training.
Table \ref{tab:probe} reports the accuracy of learning the proxy tasks from RL parameters compared to end-to-end optimization.

Most proxy tasks suffer a significant drop in accuracy ($\textgreater30\%$).
The VAE even diverges for several environments.
Learning and evaluating the decoder on the initial, random policy data does worse still, suggesting the representation is closely tuned to the current policy distribution.
Although these same proxies can improve RL, the RL representation itself seems to be different, and perhaps narrowly tuned to the task defined by reward.

\begin{table}[t]
\centering
\scalebox{0.7}{
\begin{tabular}{rrrrrrr}
\toprule
& Pong & Qbert & Seaquest & S. Invaders & BeamRider & Breakout \\
\midrule
VAE       & -      & $71\%$ & -      & $72\%$ & $65\%$ & -  \\
Reward    & $99\%$ & $63\%$ & $67\%$ & $29\%$ & $25\%$ & $44\%$ \\
Dyn. Ver. & $91\%$ & $33\%$ & $43\%$ & $38\%$ & $42\%$ & $117\%$ \\
Inv. Dyn. & $56\%$ & $62\%$ & $58\%$ & $69\%$ & $62\%$ & $81\%$ \\
\bottomrule
\end{tabular}
}
\caption{
  Analysis of proxy tasks by decoding  RL representations.
  We measure the accuracy of learning from fixed features instead of end-to-end.
  The relative performance gives some indication of what is captured by pure RL features.
  The drops in accuracy suggest that the representation is narrowly tuned to the RL task.
}
\label{tab:probe}
\end{table}

\subsection{Joint Policy and Auxiliary Optimization}
\label{sec:joint}

\begin{figure}[t]
\begin{center}
   \includegraphics[width=\linewidth]{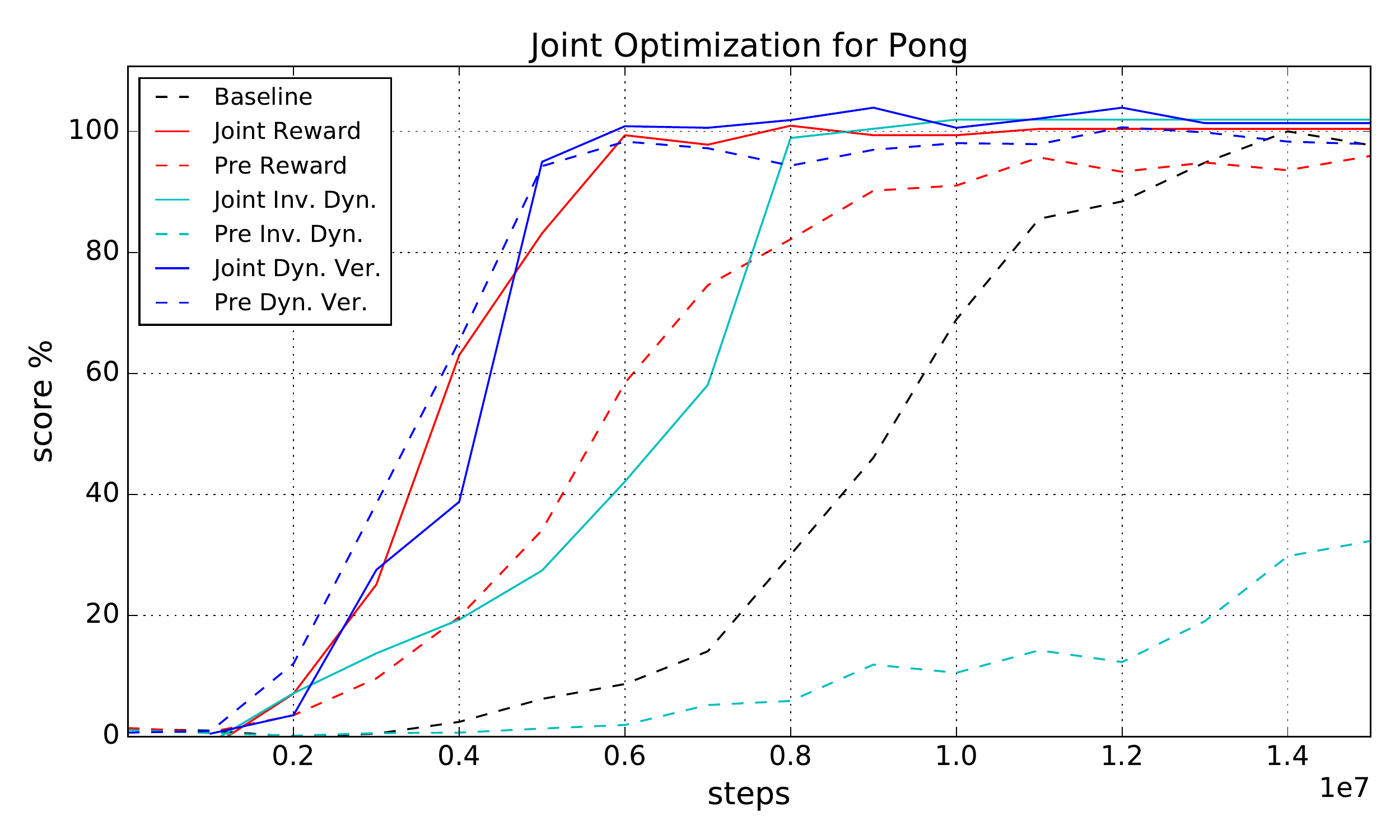}
\end{center}
\caption{
  Joint optimization with auxiliary losses further improves over pre-training.
  All of the joint tasks (solid lines) have comparable or higher data efficiency than their respective pre-trained tasks (dashed lines).
}
\label{fig:joint}
\end{figure}

Having shown that pre-training is effective in its own right, we turn to joint optimization to further boost the effects of self-supervision.
Online, multi-task optimization guarantees that the auxiliary losses are optimized on the policy distribution.
For combined supervision we simply sum the losses and gradients from reinforcement learning and self-supervision.
For a comparison of joint optimization and pre-training on Pong see Figure \ref{fig:joint}.

The joint optimization results improve on the pre-training for every task.
Note that inverse dynamics fails when pre-trained but improves over the baseline when trained jointly.
This underscores the importance of tracking the changing policy distribution: doing so helps more than any potential interference among the RL and auxiliary losses.

\subsection{Experimental Framework}

Pre-training is carried out as straightforward supervised learning on fixed data shared by all of the tasks.
The data for pre-training of each environment is collected by executing a random policy for 100,000 transitions.
The transitions collected for self-supervision are pre-processed in the same format as the transitions encountered during RL.
A portion of the collected pre-training episodes is held out for validation.

Transfer to reinforcement learning is carried out in the same manner across all tasks.
First the self-supervised output layers are discarded and replaced by outputs for policy and value.
The policy and value weights are initialized according to \citet{lecun1998efficient} as in \citet{mnih2015human}.
To control for disparities in auxiliary losses, all networks are calibrated to equalize gradients across layers by the method of \citet{krahenbuhl2015data}.
Without calibration transfer can fail to improve over random policy performance.
In rare cases, should training still fail, we fallback to transferring only the convolutional layers.

Joint optimization is carried out by summing the policy and auxiliary losses and gradients.
Each auxiliary loss has its own weight selected by cross-validation and shared across environments.
The policy loss is computed on-policy from rollouts while the auxiliary losses sample mini-batches from a small replay memory ($\textless$10,000 transitions).

For architecture we adapt the actor-critic network of \citet{mnih2016asynchronous} but reduce its capacity to that of the original DQN \citep{mnih2013playing} for computational efficiency.
For optimization we select the state-of-the art asynchronous advantage actor-critic (A3C) method \citep{mnih2016asynchronous} and configure it with comparable hyperparameters.
For the environment we adhere to the specification from DeepMind \citep{mnih2015human} by our own re-implementation through the OpenAI Gym \citep{brockman2016gym}.

Table \ref{tab:rlcheck} checks our reinforcement learning baseline against the returns of the original DQN.
Returns are better for all environments evaluated, justifying the baseline as reasonable for measuring further improvements due to self-supervision.

The code for the self-supervised tasks, policy optimization, and environment will be released.

\begin{table}[t]
\centering
\scalebox{0.68}{
\begin{tabular}{rrrrrrr}
\toprule
& Pong & Qbert & Seaquest & S. Invaders & BeamRider & Breakout \\
\midrule
NIPS DQN & 20 & 1952  & 1705 & 581  & 4092 & 168 \\
Our A3C  & 21 & 18028 & 1756 & 1102 & 5061 & 367 \\
\bottomrule
\end{tabular}
}
\caption{
  Comparison of the best scores achieved by the original DQN \citep{mnih2013playing} and the same base architecture optimized with our A3C implementation.
  Training is carried out for 60M updates.
  Scores are reported as the mean of 100 runs with random no-op starts as in existing work.
  This sanity check demonstrates reasonable returns, so improvement from self-supervision cannot be attributed to deficiencies in the RL setup.
}
\label{tab:rlcheck}
\end{table}

\section{Related Work}

Representation learning for reinforcement learning, robotics, and control is commonly known as state representation learning, as it yields the state for modeling the task as an MDP.
That is, the goal of the state representation is to transform the history of observations, actions, and rewards into a sufficient state for efficient policy learning.
This can be summarized formally as seeking a mapping $\phi$ such that the current state $s_t = \phi(o_{1:t}, a_{1:t}, r_{1:t})$ as in \citet{jonschkowski2015learning}.

Unsupervised learning by auto-encoding is a common approach to state representation learning.
The embed to control objective \citep{watter2015embed} combines variational auto-encoding with one-step dynamics modeling for image observations and locally-linear latent dynamics.
The deep spatial auto-encoder \citep{finn2016deep} maps image observations into low-dimensional spatial coordinates by auto-encoding with a smoothness prior on the latent representation.
The joint inverse and forward model of \citet{agrawal2016learning} learns to poke objects by self-supervising inverse dynamics while predicting future states (not observations) for regularization.
These approaches optimize policies to achieve a goal state without a task reward, so it is not possible to fine-tune the representation to optimize return.
In contrast our auxiliary, discriminative losses capture dynamics, inverse dynamics, and other aspects of the environment in tandem with RL.

For deep RL, the use of pre-training and transfer is limited.
ML-DDPG \citep{munk2016learning} extends actor-critic with a one-step predictive model of the successor state and reward.
The observation mapping is learned by the first layer of the the model, transferred to the actor-critic network, and then fixed.
Our successor self-supervision is discriminative rather than generative and we transfer all layers to the actor-critic network for end-to-end optimization.
The end-to-end visuomotor policies of \citet{levine2016end} have the first layer initialized from supervised pre-training on ImageNet.
Instrumented pose estimation pre-training further scaffolds the representation for policy optimization.
Our auxiliary losses are purely self-supervised and only require regular transitions.

The robotic priors of \citet{jonschkowski2015learning} are auxiliary losses for temporal coherence, repeatability, proportionality, and causality.
Multi-task optimization of these losses defines a linear, low-dimensional observation mapping for RL.
These losses are defined on distances between states conditioned on action and reward, while we define discriminative losses on the $(s, a, r, s')$ of transitions.

Concurrent work explores different methods to augment reinforcement learning with auxiliary losses.
\citet{jaderberg2016reinforcement} extend value function estimation with instantaneous reward prediction using replay memory and introduce off-policy pseudo-reward control tasks.
\citet{mirowski2016learning} extend navigation tasks with auxiliary losses for spatial and path representations through coarse depth regression and a kind of loop closure for recognizing paths that have been already visited.
\citet{dosovitskiy2016learning} learn to predict future measurements of supervised, task-specific quantities such as the presence of enemies and health in a combat game (DOOM).
In the same spirit as our work, these approaches seek to improve policy returns, data efficiency, and robustness of end-to-end RL.
Our self-supervised tasks do not require additional privileged information, we focus on discriminative formulations of auxiliary losses, and we compare a variety of ambient signals for self-supervision.

\section{Discussion}
It is encouraging that self-supervision, with and without reward, can improve reinforcement learning.
Pre-training alone suffices to show improvements especially during early iterations.
Joint training further improves data efficiency by tracking the policy distribution during optimization.

Representation learning by self-supervision alone is agnostic to any particular task, and acts as a policy scaffold no matter the reward.
This scaffold can be developed in the absence of an extrinsic reward whenever the policy is at play in the environment.
A next step is to cast these losses into intrinsic rewards to further guide optimization.
By augmenting RL with self-supervision, transitions without reward need not be so unrewarding for the representation.

\section*{Acknowledgements}

This work was supported in part by Berkeley AI Research, Berkeley Deep Drive, NSF, DARPA, NVIDIA, and Intel.
We gratefully acknowledge NVIDIA for GPU donation.
We thank John Schulman and Chelsea Finn for advice and useful discussions.
We thank Alec Radford for sharing his implementation of A3C used in our joint optimization experiments.
We thank Jeff Donahue for the care and feeding of the BiGANs.

\bibliography{rlfl}
\bibliographystyle{icml2017}

\end{document}